%% file: main_arxiv.tex
\documentclass[10pt,twocolumn,letterpaper]{article}

\usepackage{iccv}
\usepackage{times}
\usepackage{epsfig}
\usepackage{graphicx}
\usepackage{amsmath}
\usepackage{amssymb}
\usepackage{gensymb}
\usepackage{subfig}
\usepackage{float}
\usepackage{mathtools}
\usepackage{dblfloatfix}
\usepackage{mathbbol}

\usepackage{comment}
\usepackage{multirow}
\usepackage{array}
\usepackage{booktabs}

\usepackage[usenames,dvipsnames]{color}
\usepackage[table]{xcolor}    

\definecolor{fullred}{rgb}{0.85,.0,.1} 
\definecolor{navyblue}{rgb}{.0,.0,.5}
\definecolor{bleudefrance}{rgb}{0.19, 0.55, 0.91}
\definecolor{bluegray}{rgb}{0.18, 0.36, 0.6}
\definecolor{lightgray}{rgb}{0.95, 0.95, 0.95}
\definecolor{white}{rgb}{1.0, 1.0, 1.0}

\newcommand\uppersub[1]{\ensuremath{\textsf{#1}}}

\newcommand\blfootnote[1]{%
  \begingroup
  \renewcommand\thefootnote{}\footnote{#1}%
  \addtocounter{footnote}{-1}%
  \endgroup
}

\DeclareMathSymbol{\shortminus}{\mathbin}{AMSa}{"39}

\usepackage[pagebackref=true,breaklinks=true,letterpaper=true,colorlinks,bookmarks=false]{hyperref}

\iccvfinalcopy 

\def\iccvPaperID{11354} 

\ificcvfinal\fi

\begin{document}

\title{Is Pseudo-Lidar needed for Monocular 3D Object detection?}

\author{Dennis Park\thanks{equal contribution} \qquad Rareș Ambruș$^*$ \qquad Vitor Guizilini \qquad Jie Li \qquad Adrien Gaidon  \\
Toyota Research Institute \\
{\tt\small firstname.lastname@tri.global}
}

\maketitle
\ificcvfinal\thispagestyle{empty}\fi

\begin{abstract}

\input{sections/0_abstract}
\end{abstract}
\blfootnote{Code: \url{https://github.com/TRI-ML/dd3d}}

\section{Introduction}
\input{sections/1_introduction}

\section{Related work}
\input{sections/2_related_work}

\section{Dense depth pre-training for 3D detection}
\label{sec:dd3d}
\input{sections/3_dd3d}

\section{Pseudo-Lidar 3D detection}
\label{sec:pseudo_lidar}
\input{sections/4_pseudo_lidar}

\section{Experimental Setup}
\label{sec:experiments}
\input{sections/5_experiments}

\section{Results}
\label{sec:results}
\input{sections/6_results}

\vspace{0.2cm}
\section{Conclusion}
\input{sections/7_conclusion}

\clearpage
\appendix

\input{sections/8_supplementary}
\clearpage
{\small
\bibliographystyle{ieee_fullname}
\bibliography{egbib}
}

\end{document}

%% file: sections/0_abstract.tex
Recent progress in 3D object detection from single images leverages monocular depth estimation as a way to produce 3D pointclouds, turning cameras into pseudo-lidar sensors. These two-stage detectors improve with the accuracy of the intermediate depth estimation network, which can itself be improved without manual labels via large-scale self-supervised learning. However, they tend to suffer from overfitting more than end-to-end methods, are more complex, and the gap with similar lidar-based detectors remains significant.
In this work, we propose an end-to-end, single stage, monocular 3D object detector, DD3D, that can benefit from depth pre-training like pseudo-lidar methods, but without their limitations.
Our architecture is designed for effective information transfer between depth estimation and 3D detection, allowing us to scale with the amount of unlabeled pre-training data.
Our method achieves state-of-the-art results on two challenging benchmarks, with $16.34\%$ and $9.28\%$ AP for Cars and Pedestrians (respectively) on the KITTI-3D benchmark, and $41.5\%$ mAP on NuScenes.

%% file: sections/1_introduction.tex

Detecting and accurately localizing objects in 3D space is crucial for many applications, including robotics, autonomous driving, and augmented reality.
Hence, monocular 3D detection is an active research area~\cite{manhardt2019roi,simonelli2019disentangling,liu2020smoke,you2019pseudo}, owing to its potentially wide-ranging impact and the ubiquity of cameras.
Leveraging exciting recent progress in depth estimation~\cite{eigen2014depth, dorncvpr, godard2018digging2, packnet,lee2019big}, pseudo-lidar detectors~\cite{you2019pseudo,ma2020rethinking,simonelli2019disentangling} first use a pre-trained depth network to compute an intermediate pointcloud representation, which is then fed to a 3D detection network. The strength of pseudo-lidar methods is that they monotonically improve with depth estimation quality, e.g., thanks to large scale training of the depth network on raw data.

However, regressing depth from single images is inherently an ill-posed inverse problem. Consequently, errors in depth estimation account for the major part of the gap between pseudo-lidar and lidar-based detectors, a problem compounded by generalization issues that are not fully understood yet~\cite{simonelli2020demystifying}.
Simpler end-to-end monocular 3D detectors~\cite{brazil2019m3d,liu2020smoke} seem like a promising alternative, although they do not enjoy the same scalability benefits of unsupervised pre-training due to their single stage nature.

\begin{figure}[t!]
\centering
\includegraphics[width=0.47\textwidth]{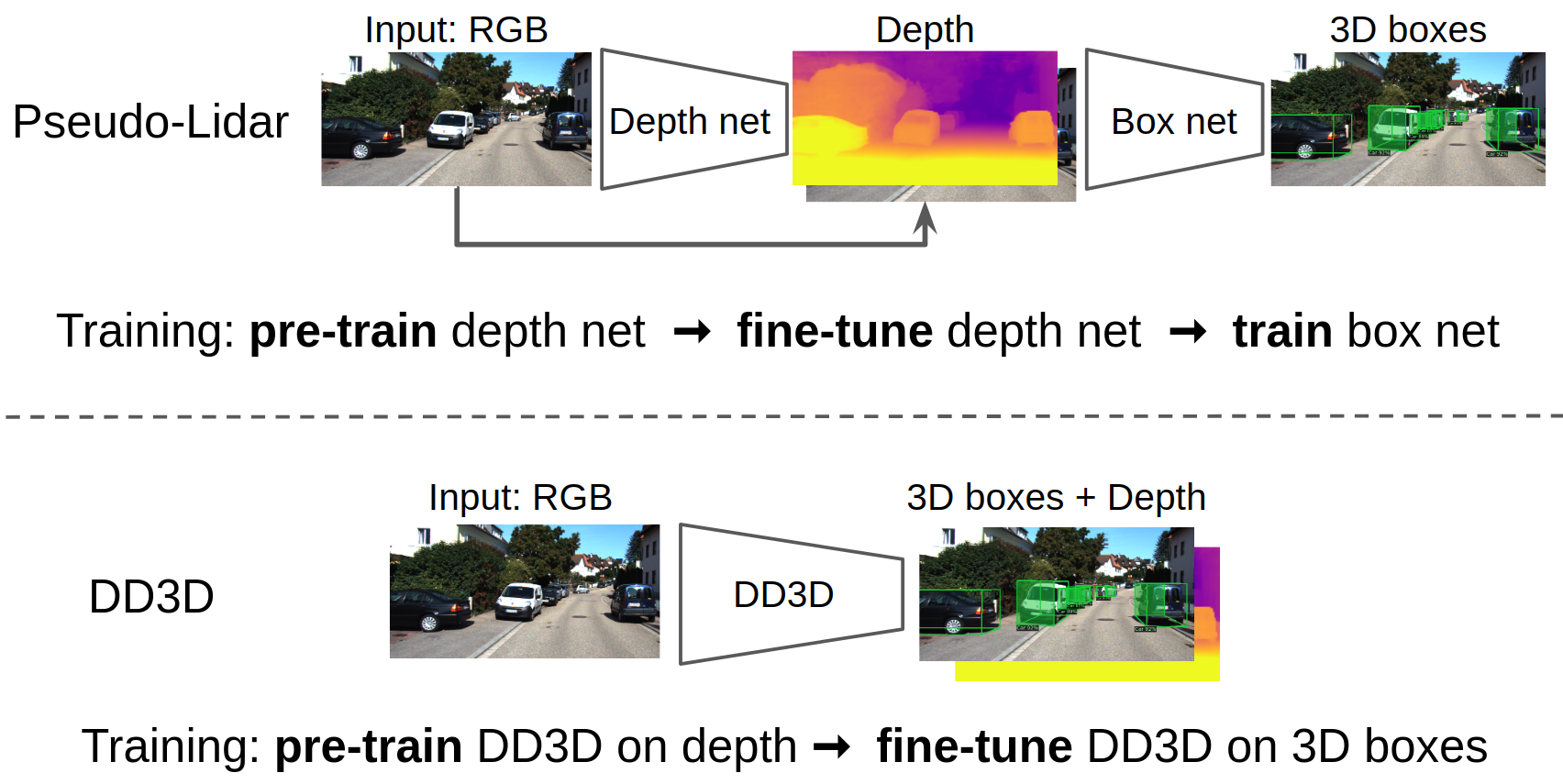}
\caption{We introduce a single-stage 3D object detector, \textbf{DD3D}, that combines the best of both pseudo-lidar methods (scaling with depth pre-training) and end-to-end methods (simplicity and generalization performance). Our detector features a simple training protocol of depth pre-training and detection fine-tuning, compared to pseudo-lidar methods that require an additional depth fine-tuning step and tend to overfit to depth errors.}
\label{fig:teaser}
\end{figure}

In this work,  we aim to get the best of both worlds: the scalability of pseudo-lidar with raw data and the simplicity and generalization performance of end-to-end 3d detectors. To this end, our \textbf{main contribution} is a novel fully convolutional single-stage 3D detection architecture, \textbf{DD3D} (for Dense Depth-pre-trained 3D Detector), that can effectively leverage monocular depth estimation for pre-training (see Figure~\ref{fig:teaser}).
Using a large corpus of unlabeled raw data, we show that DD3D scales similarly to pseudo-lidar methods, and that depth pre-training improves upon pre-training on large labeled 2D detection datasets like COCO~\cite{lin2014microsoft}, even with the same amount of data.

Our method sets a new state of the art on the task of monocular 3D detection on KITTI-3D~\cite{geiger2012we} and nuScenes~\cite{caesar2020nuscenes} with significant improvements compared to previous state-of-the-art methods. The simplicity of its training procedure  and its end-to-end optimization allows for effective use of large-scale depth data, leading to impressive multi-class detection accuracy.


%% file: sections/2_related_work.tex
\noindent\textbf{Monocular 3D detection.} A large body of work in image-based 3D detection builds upon 2D detectors, and aims to \emph{lift} them to 3D using various cues from object shapes and scene geometry. These priors are often injected by aligning 2D keypoints with their projective 3D counterparts ~\cite{chabot2017manta, ansari2018graded, barabanau2019monocular, he2019mono3dplus, qin2019monogrnet}, or by learning a low-dimensional shape representation~\cite{kundu20193drcnn, manhardt2019roi}. Other methods try to leverage geometric consistency between 2D and 3D structures, formulating the inference as a constrained optimization problem ~\cite{mousavian20173d,  liu2019fqnet, naiden2019shiftrcnn, fang20193d, choi2019mvra, kehl2017ssd, ku2019monocular, li2019gs3d}. These methods commonly use additional data (e.g. 3D CAD models, instance segmentation), or assume rigidity of objects. Inspired by Lidar-based methods \cite{chen2017mv3d, yang2018pixor, lang2019pointpillars}, another line of work employs view transformation (i.e. birds-eye-view) to overcome the limitations of perspective \emph{range} view, e.g. occlusion or variability in size \cite{kim2019bevipm, roddick2018orthographic, Srivastava2019birdgan}. These methods often require precise camera extrinsics, or are accurate only in close vicinity.

\noindent\textbf{End-to-end 3D detectors.} Alternatively, researchers have attempted to directly regress 3D bounding boxes from CNN features \cite{simonelli2019disentangling, liu2020smoke, zhou2019objects, jorgensen2019monocular, brazil2019m3d, chen2020monopair}. Typically, these approaches extend standard 2D detectors (single-stage \cite{redmon2016yolo, zhou2019objects} or  two-stage \cite{ren2015faster}) by adding \emph{heads} that predict various 3D cuboid parameterizations . The use of depth-aware convolution or dense 3D anchors \cite{ding2020learning, qin2019tlnet} enabled higher accuracy. Incorporating uncertainty in estimating the 3D box (or its depth) has also been shown to greatly improve detection accuracy~\cite{simonelli2019disentangling, chen2020monopair, simonelli2020demystifying}. In~\cite{simonelli2019disentangling}, the authors proposed a disentangled loss for 3D box regression that helps stabilize training. DD3D also falls in the end-to-end 3D detector category, however, our emphasis is on learning a good depth representation via large-scale self-supervised pre-training on raw data, which leads to robust 3D detection.



\noindent\textbf{Pseudo-Lidar methods.}
Starting with the pioneering work of~\cite{wang2019pseudo}, these methods leverage advances in monocular depth estimation and train Lidar-based detectors on the resulting pseudo pointcloud, producing impressive results~\cite{weng2019monocular, qian2020end, wang2020train,ma2019accurate}. Recent methods have improved upon~\cite{wang2019pseudo} by correcting the monocular depth with sparse Lidar readings~\cite{you2019pseudo}, decorating the pseudo pointcloud with colors~\cite{ma2019accurate}, using 2D detection to segment foreground pointcloud regions~\cite{qi2018frustum, weng2019monocular,ma2020rethinking}, and structured sparsification of the monocular pointcloud~\cite{vianney2019refinedmpl}. Recently,~\cite{simonelli2020demystifying} has shown a bias in the PL results on the representative KITTI-3D~\cite{geiger2012we} benchmark, while this paradigm remains the state-of-the-art. Multiple researchers showed that \emph{inaccurate depth estimation} is a major source of error in PL methods ~\cite{simonelli2020demystifying, wang2020foresee}. In this work, we build our reference PL method based on~\cite{qi2018frustum} and~\cite{ma2020rethinking} to investigate the benefits of using large-scale depth data and compare it with DD3D.




\noindent\textbf{Monocular depth estimation.} Estimating per-pixel depth is a key task for both DD3D (as a pre-training task) and PL (as the first of a two-stage pipeline). It is itself a thriving research area: the community has pushed toward accurate dense depth prediction via supervised ~\cite{depthcrf,huberloss,packnet-semisup,normalscvpr2,selfsupsem,packnet-semguided,fouriercvpr,dorncvpr,lee2019big} as well as self-supervised~\cite{pillai2018superdepth, zhou2018stereo, ummenhofer2017demon,packnet,godard2018digging2,vijayanarasimhan2017sfm} methods. We note that supervised monocular depth training used in this work require no annotations from human, allowing us to scale our methods to a large amount of raw data.



%% file: sections/3_dd3d.tex




Given a single image and its camera intrinsics matrix as input, the goal of monocular 3D detection is to generate a set of multi-class 3D bounding boxes relative to camera coordinates. During inference, DD3D does not require any additional data, such as per-pixel depth estimates, 2D bounding boxes, segmentations, or 3D CAD models. DD3D is also \emph{camera-aware}: it scales the depth of putative 3D boxes according to the camera intrinsics.

\subsection{Architecture}
\begin{figure}[h]
	\centering
	\includegraphics[width=1.0\columnwidth]{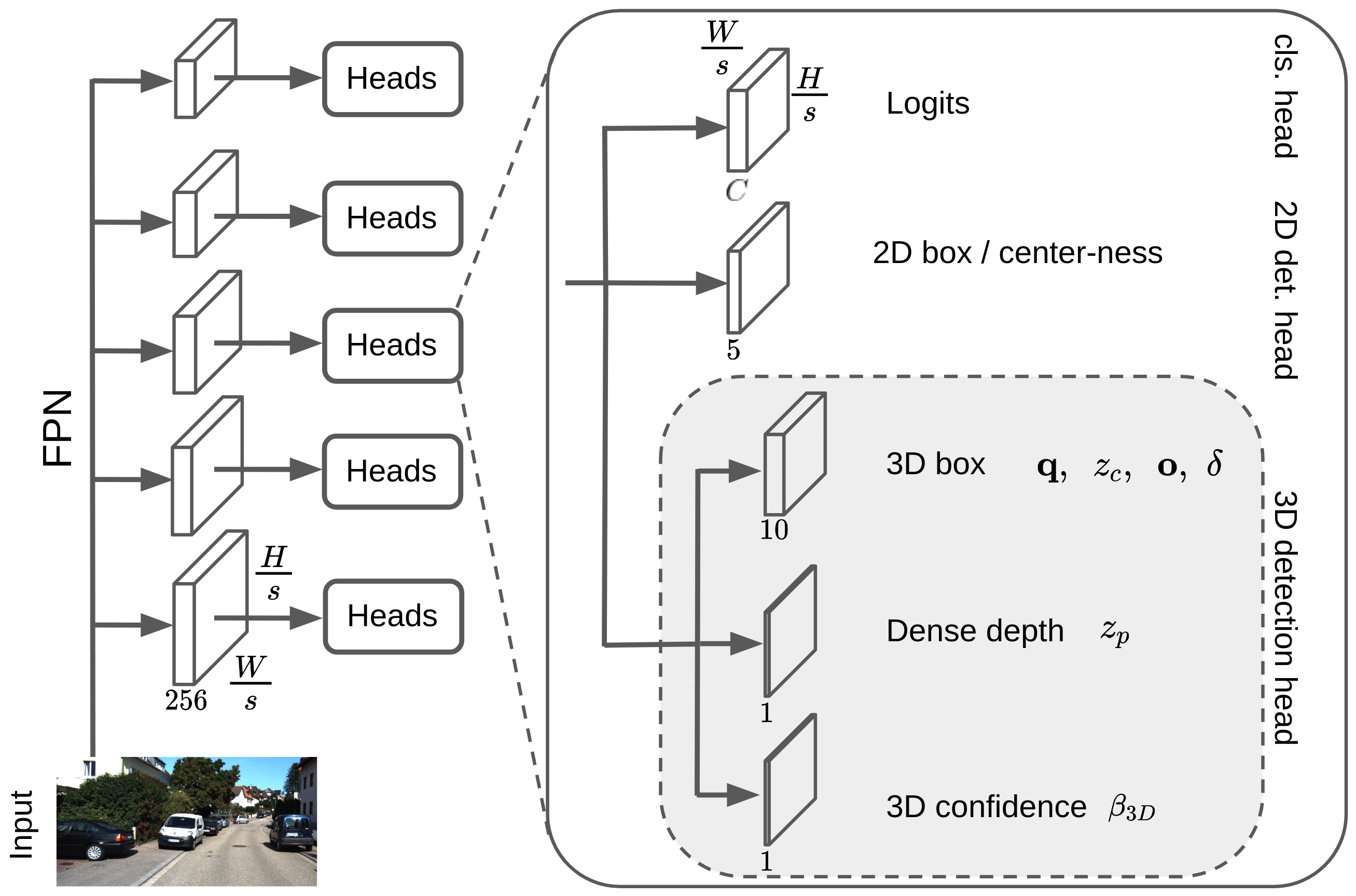}
	\caption{\textbf{DD3D} is a fully convolutional single-stage architecture that performs monocular 3D detection and dense depth prediction. To maximize reuse of pre-trained features, the inference of \emph{dense depth} and \emph{3D bounding box} share most of the parameters. The heads are shared among all FPN layers.
	}
	\label{fig:dd3d-arch}
	\vspace{-3mm}
\end{figure}

DD3D is a fully convolutional single-stage network that extends FCOS~\cite{tian2019fcos} to perform 3D detection and dense depth prediction.  The architecture (see Figure~\ref{fig:dd3d-arch}) is composed of a backbone network and three sub-networks (or \emph{head}s) that are shared among all multi-scale features. The backbone takes an RGB image as input, and computes convolutional features at different scales. As in \cite{tian2019fcos}, we adopt a feature pyramid network (FPN) \cite{lin2016fpn} as the backbone.

Three head networks are applied to each feature map produced by the backbone, and perform independent prediction tasks. 
The classification module predicts object category. It produces $C$ real values, where $C$ is the number of object categories. The 2D box module produces class-agnostic bounding boxes and center-ness by predicting $4$ offsets from the feature location to the sides of each bounding box and a scalar associated with center-ness. We refer the readers to ~\cite{tian2019fcos} for more details regarding the 2D detection architecture.

\noindent\textbf{3D detection head.} This head predicts 3D bounding boxes and per-pixel depth. It takes FPN features as input, and applies four 2D convolutions with $3 \times 3$ kernels that generate $12$ real values for each feature location. These are decoded into 3D bounding box, per-pixel depth map, and 3D prediction confidence, as described below:

\begin{itemize}
     \item $\mathbf{q} = (q_w, q_x, q_y, q_z)$ is the quaternion representation of \emph{allocentric} orientation ~\cite{manhardt2019roi, simonelli2019disentangling, liu2020smoke} of the 3D bounding box. It is normalized and transformed to an egocentric orientation \cite{manhardt2019roi}. Note that we predict orientations with the full 3 degrees of freedom. 
     
    \item $z_{\{c, p\}}$ represent the depth predictions. $z_c$ is decoded to the \emph{z}-component of 3D bounding box centers and therefore only associated with foreground features, while $z_p$ is decoded to the monocular depth to the closest surface and is associated with every pixel. To decode them to metric depth, we \emph{unnormalized} them using per-level parameters as follows:
\begin{align}
    d &= \frac{c}{p} \cdot (\sigma_l \cdot z + \mu_l), \label{eq:depth_unnorm} \\
    p & = \sqrt{{\frac{1}{f_x^2}  + \frac{1}{f_y^2}}}, 
    \label{eq:depth_decoding}
\end{align}
where $z \in \{z_c, z_p\}$ are network output, $d \in \{d_c, d_p\}$ are predicted depths, $(\sigma_l, \mu_l)$ are learnable scaling factors and offsets defined for each FPN level, $p$ is the \emph{pixel size} computed from the focal lengths, $f_x$ and $f_y$, and $c$ is a constant.
    
We note that this design of using camera focal lengths endows DD3D with \emph{camera-awareness}, by allowing us to infer the depth not only from the input image, but also from the pixel size. We found that this is particularly useful for stable training. Specifically, when the input image is resized during training, we keep the ground-truth 3D bounding box unchanged, but modify the camera \emph{resolution} as follows:
    
\begin{align}
   \mathbf{K} &= \begin{bmatrix}r_x & r_y & 1 \end{bmatrix} \begin{bmatrix}f_x & 0 & p_x \\ 0 & f_y & p_y \\ 0 & 0 & 1 \end{bmatrix},
\end{align}
   where $r_x$ and $r_y$ are resize rates, and $\mathbf{K}$ is the new camera intrinsic matrix that is used in Eqs.~\ref{eq:depth_decoding} and ~\ref{eq:center_decoding}.  Finally, $\{z_p\}$ collectively represent low-resolution versions of the dense depth maps computed from each FPN features. To recover the full resolution of the dense depth maps, we apply bilinear interpolation to match the size of the input image.
     
     \item $\mathbf{o} = (\Delta_u, \Delta_v)$ represent offsets from the feature location to the 3D bounding box center projected onto the camera plane. This is decoded to the 3D center via \emph{unprojection}:
\begin{align}
    \mathbf{C} &= \mathbf{K}^{-1} \begin{bmatrix}u_b + \alpha_l\Delta_u \\ v_b + \alpha_l\Delta_v \\ 1 \end{bmatrix} d_c,
    \label{eq:center_decoding}
\end{align}
    where $(u_b, v_b)$ is the feature location in image space, $\alpha_l$ is a learnable scaling factor assigned to each FPN level. 
    
    \item $\mathbf{\delta} = (\delta_W, \delta_H, \delta_H)$ represents the deviation in the size of the 3D bounding box from the class-specific canonical size, i.e. $\mathbf{s} = (W_0 e^{\delta_{\uppersub{W}}}, H_0 e^{\delta_{\uppersub{H}}}, D_0 e^{\delta_{\uppersub{D}}})$. As in ~\cite{simonelli2020disentangling}, $(W_0, H_0, D_0)$ is the canonical box size for each class, and is pre-computed from the training data as its average size.

    \item $\beta_{\scriptsize{\textsf{3D}}}$ represents the confidence of the 3D bounding box prediction \cite{simonelli2019disentangling}. It is transformed into a probability: $p_{\uppersub{3D}} = (1 + e^{-\beta_{\uppersub{3D}}})^{-1}$, and multiplied by the class probability computed from the classification head~\cite{tian2019fcos} to account for the relative confidence to the 2D confidence. The adjusted probability is used as the final score for the candidate.
    
\end{itemize}

\subsection{Losses}
We adopt the classification loss and 2D bounding box regresssion loss from FCOS \cite{tian2019fcos}:
\begin{equation}
\label{eq:loss_function}
\mathcal{L}_{\uppersub{2D}} = \mathcal{L}_{reg} + \mathcal{L}_{cls} + \mathcal{L}_{ctr},
\end{equation}
where the 2D box regression loss $L_{reg}$ is the IOU loss~\cite{yu2016unitbox}, the classification loss $L_{cls}$ is the binary focal loss (i.e. one-vs-all), and the center-ness loss $L_{ctr}$ is the binary cross entropy loss. For 3D bounding box regression, we use the \emph{disentangled} L1 loss described in \cite{simonelli2019disentangling}, i.e.

\begin{equation}
\label{eq:loss_function_3D}
\mathcal{L}_{\uppersub{3D}}(\mathbf B^*, \hat{\mathbf B}) = \frac{1}{8}||\mathbf B^* - \hat{\mathbf B}||_1,
\end{equation}
where $\mathbf B^*$ and $\hat{\mathbf B}$ are the $8$ vertices of ground-truth and candidate 3D boxes. This loss is replicated $4$ times by using only one of the predicted 3D box components (orientation, projected center, depth, and size), while replacing other three with their ground-truth values.

Also as in \cite{simonelli2019disentangling}, we adopt the self-supervised loss for 3D confidence which uses the error in 3D box prediction to compute a surrogate target for 3D confidence (relative to the 2D confidence): 
\begin{align}
     p^*_{\uppersub{3D}} &= e^{-\frac{1}{T} \mathcal{L}_{\uppersub{3D}}(\mathbf B^*, \hat{\mathbf B})}, 
\end{align}
where $T$ is the temperature parameter. The confidence loss $\mathcal{L}_{conf}$ is the binary cross entropy between $p_{\uppersub{3D}}$ and $p^*_{\uppersub{3D}}$. In summary, the total loss of DD3D is defined as follows:
\begin{align}
    \mathcal{L}_{\uppersub{DD}} &= \mathcal{L}_{\uppersub{2D}} + \mathcal{L}_{\uppersub{3D}} + \mathcal{L}_{conf}.
\end{align}


\subsection{Depth pre-training} 

During pre-training, we use per-pixel depth predictions from all FPN levels, i.e. $\{z_{p}\}$. We consider pixels that have valid ground-truth depth from the sparse Lidar pointclouds projected onto the camera plane, and compute L1 distance from the predicted values:
\begin{align}
    \mathcal{L}_{depth}^{l} &= ||\mathbf D^* - \hat{\mathbf D}_l||_1 \odot \mathcal{\mathbf M}, \label{eq:dense_depth_loss} \\
    \mathcal{L}_{depth} &= \sum_l \mathcal{L}_{depth}^l,
\end{align}
where $\mathbf D^*$ is the ground-truth depth map, $\hat{\mathbf D_l}$ is the predicted depth map from the $l$-th level in FPN (i.e. interpolated $z_p$), and $\mathcal{\mathbf M}$ is the binary indicator for valid pixels. We observed using all FPN levels in the objective, rather than e.g. using only the highest resolution features, enables stable training, especially when training \emph{from scratch}. We also observed that the L1 loss yields stable training with large batch sizes and high-resolution input, compared to SILog loss~\cite{eigen2014depth,lee2019big} that is popular in monocular depth estimation literature.

We note that the two paths in DD3D from the input image to the 3D bounding box and to the dense depth prediction differ only in the last $3 \times 3$ convolutional layer, and thus share nearly all parameters. This allows for effective transfer from the pre-trained representation to the target task.

While pre-training, the camera-awareness of DD3D allows us to use camera intrinsics that are substantially different from the ones of the target domain, while still enjoying effective transfer. Specifically, from Eqs.~\ref{eq:depth_unnorm} and ~\ref{eq:depth_decoding}, the error in depth prediction, $z$, caused by the difference in resolution between the two domains is corrected by the difference in \emph{pixel size}, $p$.

%% file: sections/4_pseudo_lidar.tex
An alternative way to utilize a large set of image-Lidar frames is to adopt the \textit{Pseudo-Lidar (PL)} paradigm and aim to improve the depth network component using the large scale data. PL is a two-stage method: first, given an image it applies a monocular depth network to predict per-pixel depth. The dense depth map is transformed into a 3D point cloud, and then a Lidar-based 3D detector is used to predict 3D bounding boxes. The modularity of PL methods enables us to quantify the role of improved depth predictors brought by a large-scale image-LiDAR dataset (see the supplementary material for additional details.)



\noindent\textbf{Monocular depth estimation.} The aim of monocular depth estimation is to compute the depth $\hat{D} = f_D\left( I\left(p\right)\right)$ for each pixel $p \in I$. Similarly to Eq.~\ref{eq:dense_depth_loss}, given the ground truth depth measurement $\mathbf{D^*}$ acquired from Lidar pointclouds, we define a loss by the error between the predicted and the ground-truth depth. Here we instead use the SILog loss~\cite{eigen2014depth,lee2019big}, which yields better performance than L1 for PackNet.

\noindent\textbf{Network architecture.} As the depth network, we use PackNet~\cite{guizilini20203d}, a state-of-the-art monocular depth prediction architecture which uses packing and unpacking blocks with 3D convolutions. By avoiding feature sub-sampling, PackNet recovers fine structures in the depth map with high precision; moreover, PackNet has been shown to generalize better thanks to its increased capacity~\cite{guizilini20203d}.







\noindent\textbf{3D detection.} To predict 3D bounding boxes from the input image and the estimated depth map, we follow the method proposed by~\cite{qi2018frustum,ma2020rethinking}. We first convert the estimated depth map into a 3D pointcloud similarly to Eq.~\ref{eq:center_decoding}, and concatenate each 3D point with the corresponding pixel values. This results in a 6D tensor encompassing colors along with 3D coordinates. We use an off-the-shelf 2D detector to identify proposal regions in input images, and apply a 3D detection network to each RoI region of the 6-channel image to produce 3D bounding boxes.

\noindent{\textbf{Backbone, detection head and 3D confidence.}}
We follow~\cite{ma2020rethinking} and process each RoI with a ResNet-18~\cite{he2016deep} backbone that uses Squeeze-and-Excitation layers~\cite{hu2018squeeze}. As the RoI contains both objects as well as background pixels, the resulting features are filtered via foreground masks computed based on the associated RoI depth map~\cite{ma2019accurate}. The detection head follows~\cite{ma2020rethinking} and operates in 3 distance ranges, producing one bounding box for each range. The final output is then selected based on the mean depth of the input RoI. Following~\cite{simonelli2019disentangling,simonelli2020demystifying}, we modify the detection head to also output a 3D confidence value $\gamma$ per detection, which is linked to the 3D detection loss. 

\noindent{\textbf{Loss function.}} The 3D regression loss~\cite{qi2018frustum} is defined as:
\begin{equation} \label{eq:3D_bbox_loss}
\small
\mathcal{L}_\uppersub{3D}^\uppersub{PL} = \mathcal{L}_{center} +  \mathcal{L}_{size} +  \mathcal{L}_{heading} +  \mathcal{L}_{corners}.
\end{equation}
In addition, we define a loss that links the predicted 3D confidence $\gamma$ with the 3D bounding box coordinates loss~\cite{simonelli2019disentangling} using a Binary Cross Entropy (BCE) formulation with target $\hat{\gamma}=e^{-\mathcal{L}_{corners}}$. The final PL 3D detection loss is: $\mathcal{L}_\uppersub{PL} = \mathcal{L}_\uppersub{3D}^\uppersub{PL} +  \mathcal{L}_{conf}^\uppersub{PL}.$





%% file: sections/5_experiments.tex
\subsection{Datasets}
\label{sec:datasets}
\input{sections/5_1_datasets}

\subsection{Implementation detail}
\label{sec:implementation}
\input{sections/5_2_implementation}

%% file: sections/5_1_datasets.tex
\noindent\textbf{KITTI-3D.} The KITTI-3D detection benchmark~\cite{geiger2012we} consists of urban driving scenes with $8$ object classes. The benchmark evaluates 3D detection accuracy on three classes (Car, Pedestrian, and Cyclist) using two average precision (AP) metrics computed with class-specific thresholds on intersection-over-union (IoU) of 3D bounding boxes or Bird-Eye-View (2D) bounding boxes. We refer to these metrics as \emph{3D AP} and \emph{BEV AP}. We use the revised AP$|_{R_{40}}$ metrics~\cite{simonelli2020disentangling}. The training set consists of $7481$ images, the test set of $7518$ images. The objects in the test set are organized into three partitions according to their difficulty level (easy, moderate, hard), and are evaluated separately. For the analysis in Section~\ref{subsec:ablative}, we follow the common practice of splitting the training set into $3712$ and $3769$ images~\cite{chen2015monocular}, and report validation results on the latter. We refer to these splits as KITTI-3D \emph{train} and KITTI-3D \emph{val}.

\noindent\textbf{nuScenes.} The nuScenes 3D detection benchmark~\cite{caesar2020nuscenes} consists of $1000$ multi-modal videos with $6$ cameras covering the full $360$-degree field of view. The videos are split into $700$ for training, $150$ for validation, and $150$ for testing. The benchmark requires to report 3D bounding boxes of $10$ object classes over 2Hz-sampled video frames. The evaluation metric, \emph{nuScenes detection score} (\emph{NDS}), is computed by combining the detection accuracy (\emph{mAP}) computed over four different thresholds on center distance with five \emph{true-positive} metrics. We  report NDS and mAP, along with the three true-positive metrics that concern 3D detection, i.e. \emph{ATE}, \emph{ASE}, and \emph{AOE}.


\noindent\textbf{KITTI-Depth.} We use the KITTI-Depth dataset~\cite{geiger2012we} to fine-tune the depth networks of our PL models. It contains over $93$ thousands depth maps associated with the images in the KITTI \emph{raw} dataset. The standard monocular depth protocol~\cite{zhou2017unsupervised,godard2018digging2,packnet} is to use the \emph{Eigen} splits~\cite{eigen2014depth}. However, as described in~\cite{simonelli2020demystifying}, up to a third of its training images overlap with KITTI-3D images, leading to biased results for PL models. To avoid this bias, we generate a new split by removing training images that are geographically close (i.e. within $50$m) to any of the KITTI-3D images. We denote this split by \textit{Eigen-clean} and use it to fine-tune the depth networks of our PL models.  

\noindent\textbf{DDAD15M.} To pre-train DD3D and our PL models, we use an in-house dataset that consists of $25000$ multi-camera videos of urban driving scenes. DDAD15M is a larger version of DDAD~\cite{packnet}: it contains high-resolution Lidar sensors to generate pointclouds and $6$ cameras synchronized with $10$ Hz scans. Most videos are 10-second long, which amounts to approximately $15$M image frames in total. Unless noted differently, we use the entire dataset for pre-training.

%% file: sections/5_2_implementation.tex
\textbf{DD3D.} We use V2-99~\cite{lee2019centermask}  extended to an FPN as the backbone network. When pre-training DD3D, we first initialize the backbone with parameters pre-trained on the 2D detection task using the COCO dataset~\cite{lin2014microsoft}, and  perform the pre-training on dense depth prediction using the DDAD15M dataset. 

We use a test-time augmentation by resizing and flipping the input images. We observed $2.3\%$ gain in ``Car'' BEV AP on KITTI \emph{val}, but no improvement on the nuScenes validation set. All metrics on DD3D in Section~\ref{subsec:ablative} are averages over $4$ training runs. We observed the variance over the runs to be small, i.e. $0.5\sim1.2\%$ BEV AP. 



\textbf{PL.} When training PackNet~\cite{packnet}, we use only the front camera images of DDAD15M to pre-train PackNet, and train until convergence. We then fine-tune the network on KITTI \textit{Eigen-clean} split for $5$ epochs. For more details on training DD3D and PL, please refer to the supplementary material.



%% file: sections/6_results.tex
\subsection{Benchmark evaluation}
\label{subsec:benchmark}
\input{sections/6_1_results_benchmark}

\subsection{Analysis}
\label{subsec:ablative}
\input{sections/6_2_results_ablative}

%% file: sections/6_1_results_benchmark.tex
In this section, we evaluate DD3D on the KITTI-3D and nuScenes benchmarks. DD3D is pre-trained on DDAD15M, and then fine-tuned on the training set of each dataset. We also evaluate PL on KITTI-3D. Its depth network is pre-trained on DDAD15M and fine-tuned on KITTI \emph{Eigen-clean}, and its detection network is trained on the KITTI-3D \emph{train}. 



\input{tables/kitti3d_test} 
\input{tables/nuscenes_summary}
\input{tables/kitti_val_ablation_v3} 
\input{tables/kitti3d_test_multi_2} 
\input{tables/nuscenes_results}

\begin{figure*}[h]
\vspace{-2mm}
\centering
\includegraphics[width=0.49\textwidth]{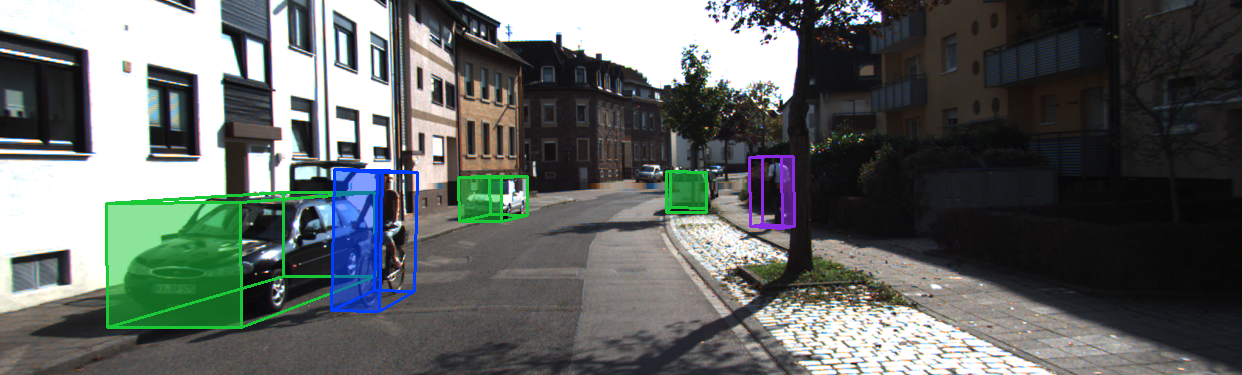}
\includegraphics[width=0.49\textwidth]{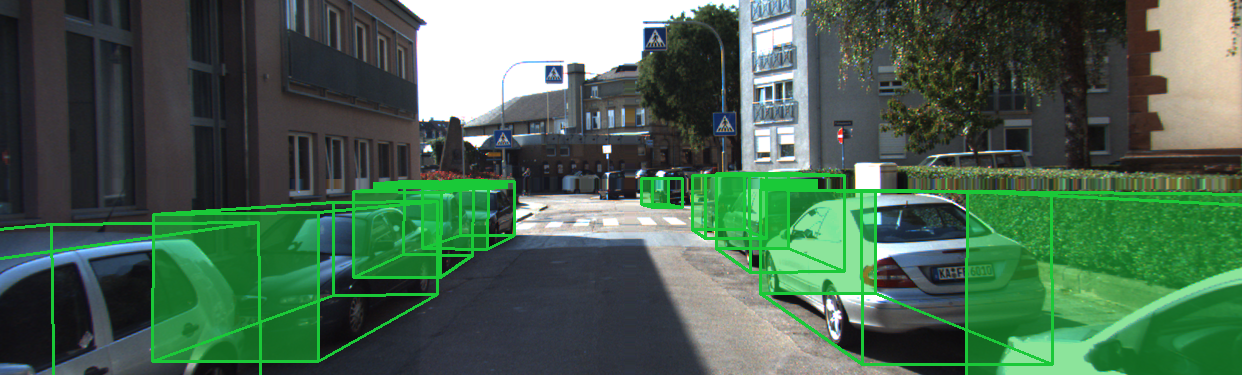} \\
\includegraphics[width=0.49\textwidth]{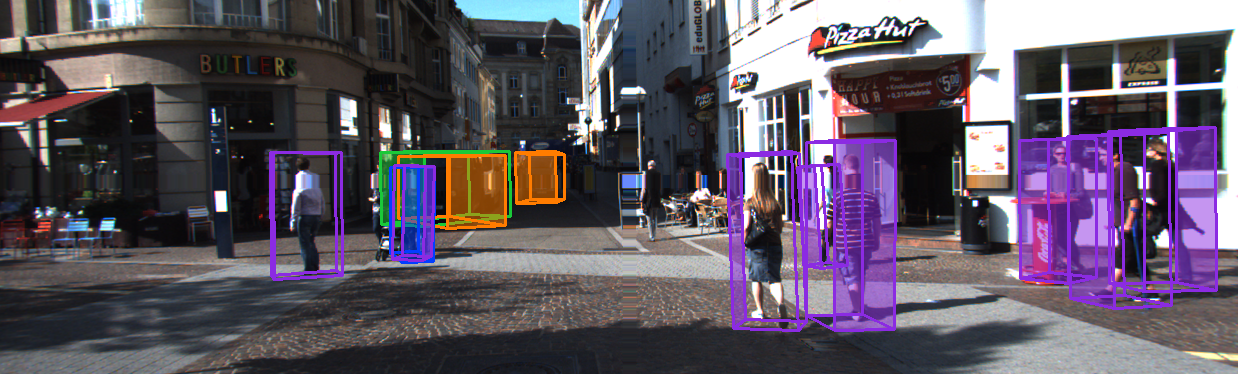}
\includegraphics[width=0.49\textwidth]{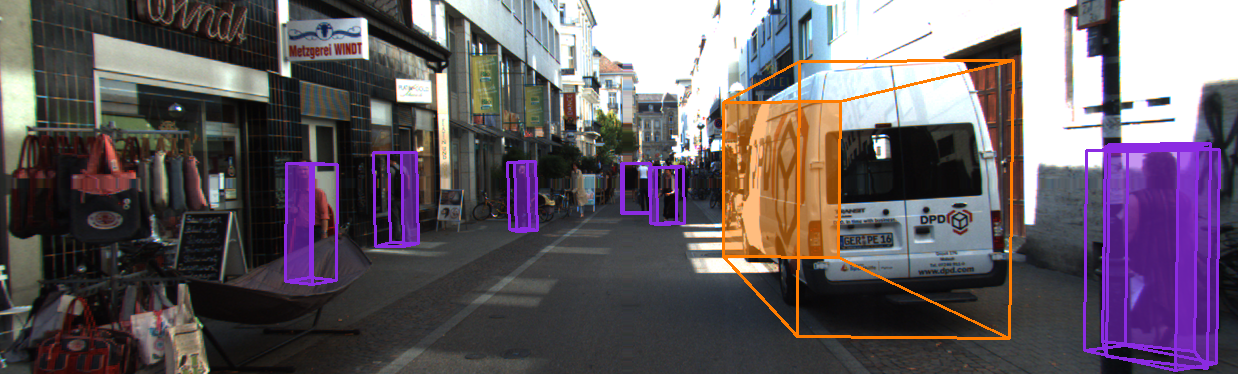} \\
\includegraphics[width=0.325\textwidth]{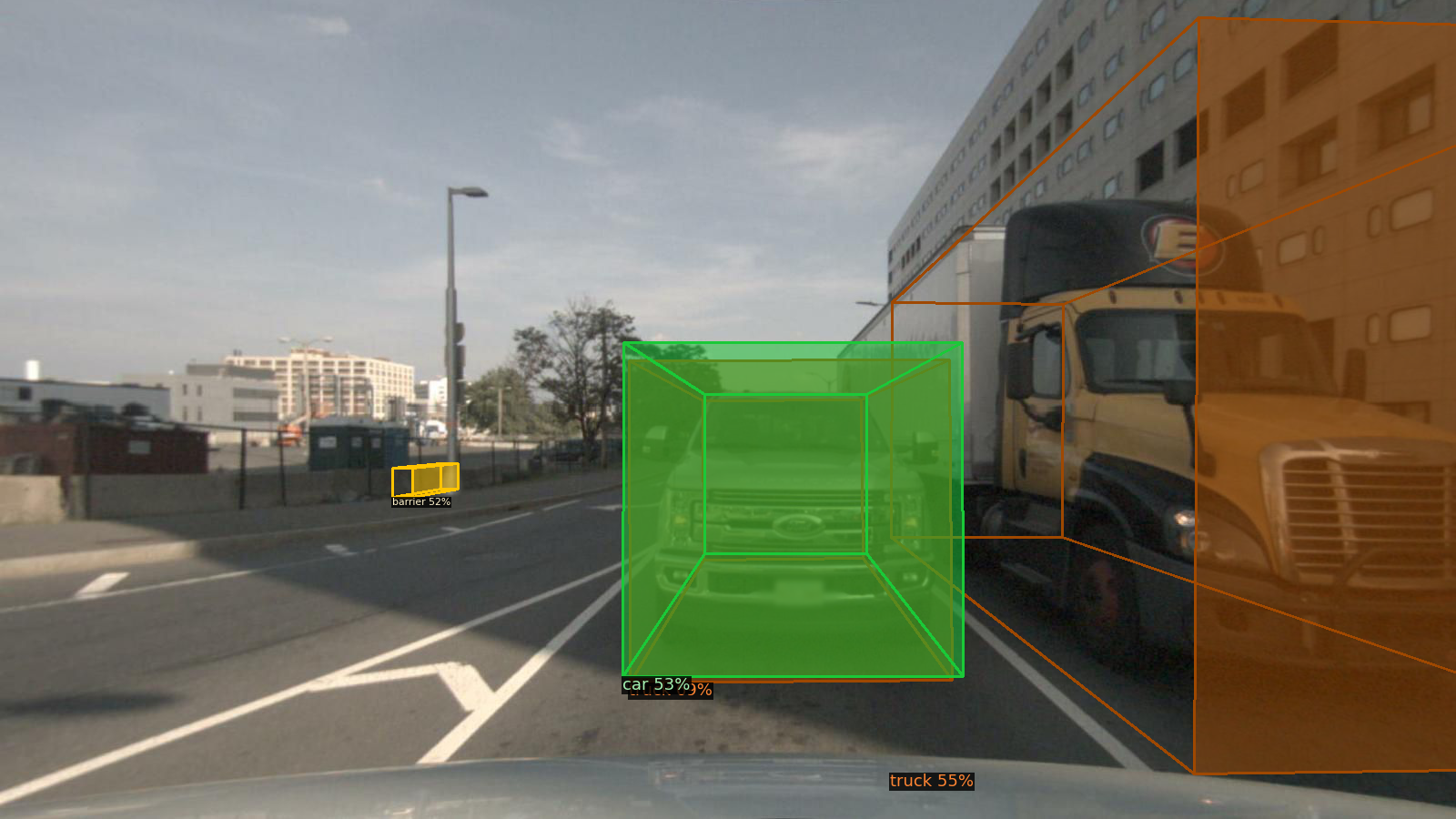}
\includegraphics[width=0.325\textwidth]{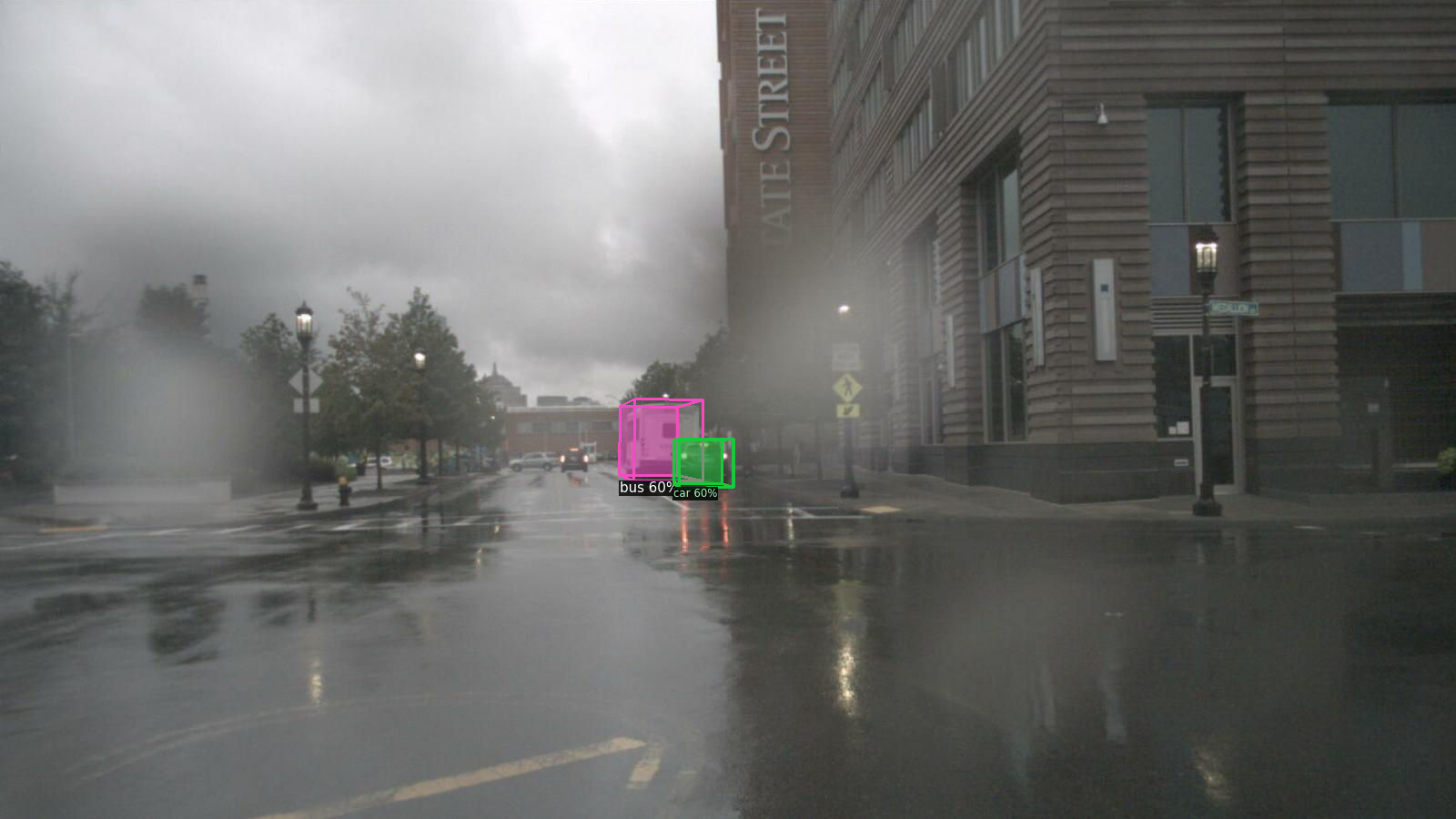}
\includegraphics[width=0.325\textwidth]{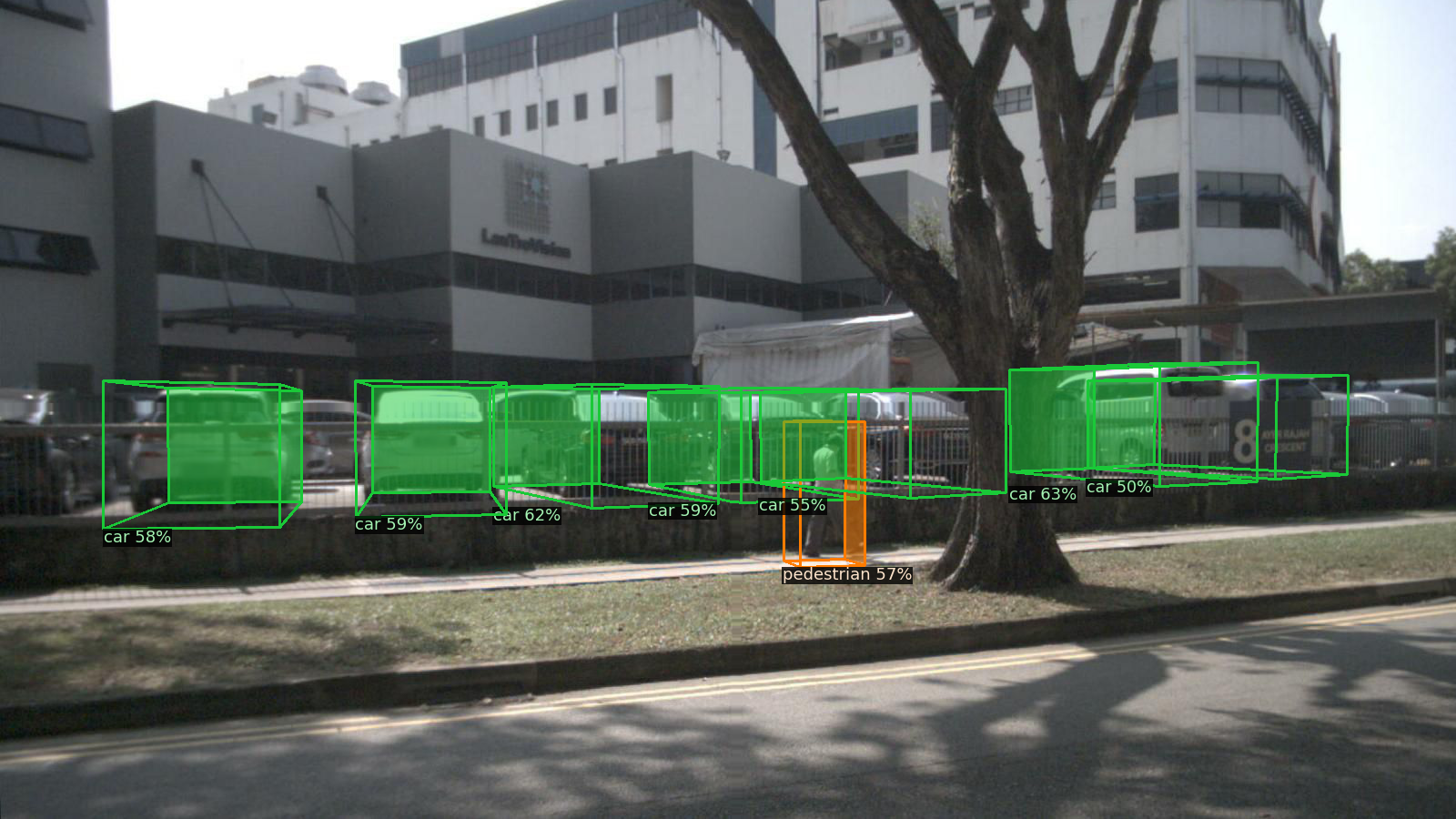} 

\vspace*{-2mm}
\caption{
\textbf{Qualitative visualization of DD3D detections.} The first two rows are from the KITTI-3D dataset and the last row is from nuScenes. None of the images were seen during training. 
}
\label{fig:results_qualitative}
\end{figure*}

\textbf{KITTI-3D.} In Table~\ref{table:kitti_3d_test} we compare the accuracy of DD3D to state-of-the-art methods on the KITTI-3D benchmark. DD3D achieves a significant improvement over all methods with $\mathbf{16.34\%}$ 3D AP on \emph{Moderate Cars}, which amounts to a $\mathbf{23\%}$ improvement from the previous best method ($13.25\%$). Qualitative visualization is presented in Figure~\ref{fig:results_qualitative}. In Table~\ref{table:kitti_3d_test_multi_class} we  evaluate DD3D also on the \textit{Pedestrian} and \textit{Cyclist} classes. DD3D outperforms all other approaches on the \textit{Pedestrian} category, with an $\mathbf{80.5\%}$ improvement from the previous best method ($\mathbf{9.30\%}$ vs $5.14\%$ 3D AP). On \textit{Cyclist} DD3D achieves the second best result, reducing the gap to~\cite{ku2019monocular} that uses ground-truth pointclouds to train a per-instance pointcloud reconstruction module and a two-stage regression network to refine 3D object proposals.  


We next report the accuracy of our PL detector in Table~\ref{table:kitti_3d_test}.  The accuracy of our PL detector is on par with state-of-the-art methods, however it performs significantly worse than DD3D ($13.05\%$ vs. $16.34\%$). We will discuss this result in the context of generalizability of PL methods in Section~\ref{subsec:ablative}.



\noindent\textbf{nuScenes.} In Table~\ref{table:nuscenes_test_summary} we compare DD3D with other monocular methods reported on the nuScenes detection benchmark. The metrics are averages over all $10$ categories of the dataset. DD3D outperforms all other methods with a $\mathbf{17\%}$  improvement in mAP compared to the previously best published method~\cite{wang2021fcos3d} ($\mathbf{41.8\%}$ vs. $35.8\%$ mAP) as well as a $13\%$ improvement compared to the best (unpublished) method. We note that DD3D even surpasses PointPillars~\cite{lang2019pointpillars}, which is a Lidar based detector. 

In Table~\ref{table:nuscenes_test_detailed} we compare per-class accuracy of DD3D with other methods on the three major categories, with various thresholds on distance. In general, DD3D offers significant improvements across all categories and thresholds. In particular, DD3D performs significantly better on the stricter criteria: comparing to the previous best method \cite{wang2021fcos3d}, the relative improvements averaged across the 3 object classes are $\mathbf{103.7\%}$ and $\mathbf{35.5\%}$ for $0.5$m and $1.0$m thresholds, respectively.

%% file: tables/kitti3d_test.tex
\begin{table}[t!]
\centering
{
\footnotesize
\setlength{\tabcolsep}{0.4em}
\rowcolors{2}{lightgray}{white}
\begin{tabular}{l|ccc|ccc}
\toprule
& \multicolumn{6}{c}{Car} \\
& \multicolumn{3}{c}{BEV AP} & \multicolumn{3}{c}{3D AP} \\ 
\multirow{-2}{*}{Methods}& 
Easy & 
Med &
Hard &
Easy & Med & Hard \vspace{0.5mm}\\
\midrule


ROI-10D~\cite{manhardt2019roi} & 
9.78 & 
4.91 &
3.74 &
4.32 &
2.02 &
1.46
\\
GS3D~\cite{li2019gs3d} & 
8.41 & 
6.08 &
4.94 &
4.47 &
2.90 &
2.47
\\
MonoGRNet~\cite{qin2019monogrnet} & 
18.19 & 
11.17 &
8.73 &
9.61 &
5.74 &
4.25
\\
MonoPSR~\cite{ku2019monocular} & 
18.33 & 
12.58 &
9.91 &
10.76 &
7.25 &
5.85
\\

MonoPL~\cite{weng2019monocular} & 
-  & 
-  &
-  &
10.76 &
7.50 &
6.10
\\

SS3D~\cite{jorgensen2019monocular} & 
16.33 & 
11.52 &
9.93 &
10.78 &
7.68 &
6.51
\\

MonoDIS (single)~\cite{simonelli2019disentangling} & 
17.23 & 
13.19 &
11.12 &
10.37 &
7.94 &
6.40
\\

M3D-RPN~\cite{brazil2019m3d} & 
21.02 & 
13.67 &
10.23 &
14.76 &
9.71 &
7.42
\\

SMOKE~\cite{liu2020smoke} & 
20.83 & 
14.49 &
12.75 &
14.03 &
9.76 &
7.84
\\

MonoPair~\cite{chen2020monopair} & 
19.28 & 
14.83 &
12.89 &
13.04 &
9.99 &
8.65
\\

AM3D~\cite{ma2019accurate} & 
25.03 & 
17.32 &
14.91 &
16.50 &
10.74 &
9.52
\\

PatchNet ~\cite{ma2020rethinking} & 
22.97 & 
16.86 &
14.97 &
15.68 &
11.12 &
10.17
\\

RefinedMPL~\cite{vianney2019refinedmpl} & 
28.08 & 
17.60 &
13.95 &
18.09 &
11.14 &
8.96
\\

D4LCN~\cite{ding2020learning} & 
22.51 & 
16.02 &
12.55 &
16.65 &
11.72 &
9.51
\\

Kinematic3D~\cite{brazil2020kinematic} & 
26.99 & 
17.52 &
13.10 &
19.07 &
12.72 &
9.17
\\

MonoDIS (multi)~\cite{simonelli2020disentangling} & 
24.45 & 
\underline{19.25} &
\underline{16.87} &
16.54 &
12.97 &
11.04
\\

Demystifying~\cite{simonelli2020demystifying} &
- & 
- &
- &
\textbf{23.66} &
\underline{13.25} &
\underline{11.23}
\\

\midrule

PL & 
\underline{28.87} & 
18.57 &
15.74 &
20.78 &
13.05 &
10.66
\\

DD3D & 
\textbf{30.98} & 
\textbf{22.56} &
\textbf{20.03} &
\underline{23.22} &
\textbf{16.34} &
\textbf{14.20}
\\


\bottomrule
\end{tabular}\\\vspace{0mm}
\caption{
\textbf{KITTI-3D \textit{test} set evaluation on \emph{Car}.} We report AP$|_{R_{40}}$ metrics. \textbf{Bold} and \underline{underline} denote the best and second best results.
}
\label{table:kitti_3d_test}
}
\end{table}

%% file: tables/nuscenes_summary.tex



\begin{table}[h]
\centering
{
\tiny
\footnotesize
\setlength{\tabcolsep}{0.3em}
\rowcolors{2}{lightgray}{white}
\begin{tabular}{l|c|c|c|c|c}
\toprule
Metric & AP[\%]$\uparrow$ & ATE[m]$\downarrow$  & ASE[1-IoU]$\downarrow$ & AOE[rad]$\downarrow$ &  NDS$\uparrow$  \\
\midrule

CenterNet$^*$                          & 33.8 & 0.66 & 0.26 & 0.63 & 0.40 \\ 
AIML-ADL$^*$                           & 35.2 & 0.70 & 0.26 & 0.39 & 0.43 \\
DHNet$^*$                              & 36.3 & 0.67 & 0.26 & 0.40 & 0.44 \\
PGDepth$^*$                            & 37.0 & 0.66 & 0.25 & 0.49 & 0.43 \\
P.Pillars~\cite{lang2019pointpillars}       & 31.0 & \textbf{0.52} & 0.29 & 0.50 & \underline{0.45} \\
MonoDIS~\cite{simonelli2020disentangling}   & 30.4 & 0.74 & 0.26 & 0.55 & 0.38 \\
FCOS3D~\cite{wang2021fcos3d}                & \underline{35.8} & 0.69 & \textbf{0.25} & \underline{0.45} & 0.43 \\
\midrule
DD3D                                        & \textbf{41.8} & \underline{0.57} & \textbf{0.25} &\textbf{ 0.37} & \textbf{0.48} \\


 
\end{tabular}\\\vspace{0mm}
\caption{
\textbf{nuScenes detection \emph{test} set evaluation.} We present summary metrics of the benchmark. * denotes results reported on the benchmark that do not have associated publications at the time of writing. The \underline{underline} denotes the second best published approach. Note that PointPillars~\cite{lang2019pointpillars} is a Lidar-based detector.
}
\label{table:nuscenes_test_summary}
}
\end{table}

%% file: tables/kitti_val_ablation_v3.tex
\begin{table}[t!]
\centering
{
\footnotesize
\setlength{\tabcolsep}{1.0em}
\rowcolors{2}{lightgray}{white}
\begin{tabular}{l|ccc}
\toprule
& \multicolumn{3}{c}{BEV AP} \\ 
\multirow{-2}{*}{Methods}& 
Easy & 
Mod &
Hard \\

\midrule


MonoDIS (single)~\cite{simonelli2019disentangling} & 
18.5 & 
12.6 &
10.7
\\

M3D-RPN~\cite{brazil2019m3d} & 
20.9 & 
15.6 &
11.9 
\\



Kinematic3D~\cite{brazil2020kinematic} & 
27.8 & 
19.7 &
15.1
\\

\midrule
\multicolumn{4}{c}{DD3D with DLA-34 backbone} \\ 
\midrule

DD3D \scriptsize{(with DLA-34)} &
33.5 &
26.0 &
22.6 
\\

\quad {\large -} DDAD15M & 
26.8 &
20.2 &
16.7 
\\

\quad {\large -} COCO &
31.7 &
24.0 &
20.3
\\


\midrule
\multicolumn{4}{c}{DD3D with V2-99 backbone} \\ 

\midrule

DD3D \scriptsize{(with V2-99)} &
37.0 &
29.4 &
25.4
\\

\quad {\large -} DDAD15M &
25.5 &
18.7  &
15.5 
\\

\midrule
\multicolumn{4}{c}{Pseudo-Lidar} \\ 
\midrule

PL (\scriptsize{DDAD15M $\rightarrow$ KITTI}) &  
43.5 & 
30.1 &
25.4 
\\

\quad {\large -} KITTI &  
25.8 & 
19.1 &
16.4
\\

\quad {\large -} DDAD15M &  
27.6 &	
19.2 &	
16.4
\\

\bottomrule
\end{tabular}\\\vspace{0mm}
\caption{
\textbf{Ablative analysis of DD3D and PL on the KITTI-3D \textit{validation} set.} As pre-training methods, KITTI denotes dense depth prediction using \emph{Eigen-clean} split of KITTI-Depth dataset, DDAD15M denotes dense depth prediction using DDAD15M dataset, and COCO denotes initial pre-training phase on 2D detection. The right arrow ($\rightarrow$) denotes sequential pre-training phases. We report BEV AP$|_{R_{40}}$ metrics on \emph{Car}. The analysis is presented in Section~\ref{subsec:ablative}.
\label{table:kitti_3d_val_ablation}
}
}
\end{table}


%% file: tables/kitti3d_test_multi_2.tex
\begin{table*}[t]
\centering
{
\footnotesize
\setlength{\tabcolsep}{1em}
\rowcolors{2}{lightgray}{white}
\begin{tabular}{l|ccc|ccc|ccc|ccc}
\toprule
& \multicolumn{6}{c}{Pedestrian} & \multicolumn{6}{c}{Cyclist} \\
& \multicolumn{3}{c}{BEV AP} & \multicolumn{3}{c}{3D AP} & \multicolumn{3}{c}{BEV AP} & \multicolumn{3}{c}{3D AP} \\ 
\multirow{-3}{*}{Methods}& 
Easy & 
Med &
Hard &
Easy & Med & Hard & Easy & 
Med &
Hard &
Easy & Med & Hard\vspace{0.5mm}\\
\midrule
OFTNet~\cite{roddick2018orthographic} &
1.28 &
0.81 &
0.51 &
0.63 &
0.36 &
0.35 &
0.36 &
0.16 &
0.15 &
0.36 &
0.16 &
0.15 \\

SSD3D~\cite{jorgensen2019monocular} &
2.48 &
2.09 &
1.61 &
2.31 &
1.78 &
1.48 &
\underline{3.45} &
1.89 &
1.44 \\

M3D-RPN~\cite{brazil2019m3d} &
5.65 &
4.05 &
3.29 &
4.92 &
3.48 &
2.94 &
1.25 &
0.81 &
0.78 &
0.94 &
0.65 &
0.47 \\

MonoPSR~\cite{ku2019monocular} &
7.24 &
4.56 &
4.11 &
6.12 &
4.00 &
3.30 &
\textbf{9.87} &
\textbf{5.78} &
\textbf{4.57} &
\textbf{8.37} &
\textbf{4.74} &
\textbf{3.68} \\

MonoDis (multi)~\cite{simonelli2020disentangling} &
\underline{9.07} &
\underline{5.81} &
\underline{5.09} &
\underline{7.79} &
\underline{5.14} &
\underline{4.42} &
1.47 &
0.85 &
0.61 &
1.17 &
0.54 &
0.48 \\

\midrule

DD3D &
\textbf{15.90} &
\textbf{10.85} &
\textbf{8.05} &
\textbf{13.91} &
\textbf{9.30} &
\textbf{8.05} &
3.20 &
\underline{1.99} &
\underline{1.79} &
\underline{2.39} &
\underline{1.52} &
\underline{1.31} \\

\bottomrule
\end{tabular}\\\vspace{0mm}
\caption{
\textbf{KITTI-3D \textit{test} set \emph{Pedestrian} and \emph{Cyclist} results.}
}
\label{table:kitti_3d_test_multi_class}
}
\end{table*}


%% file: tables/nuscenes_results.tex
\begin{table*}
\centering
{
\footnotesize
\setlength{\tabcolsep}{1em}
\rowcolors{2}{lightgray}{white}
\begin{tabular}{l|cccc|cccc|cccc}
\toprule
& \multicolumn{4}{c}{Car [\%] $\uparrow$} & \multicolumn{4}{c}{Pedestrian [\%] $\uparrow$} &  \multicolumn{4}{c}{Bicycle [\%] $\uparrow$} \\

\multirow{-2}{*}{Methods}& 
0.5m & 1.0m & 2.0m & 4.0m & 0.5m & 1.0m & 2.0m & 4.0m & 0.5m & 1.0m & 2.0m & 4.0m\vspace{0.5mm}\\

\midrule

CenterNet$^*$  &
20.0 & 45.8 & 68.0 & 80.6 &
7.9 & 26.7 & 49.6 & 65.9 &
4.3 & 13.8 & 28.4 & 36.2 \\

AIML-ADL$^*$ &
14.2 & 36.8 & 58.5 & 71.0 &
9.6 & 30.8 & 54.6 & 69.4 &
5.2 & 21.6 & 37.3 & 46.2 \\

DHNet$^*$ &
15.2 & 37.9 & 59.4 & 71.5 &
10.5 & 31.7 & 55.7 & 69.9 &
5.6 & 24.2 & 38.9 & 48.0 \\

PGDepth$^*$    &
17.0 & 43.6 & 67.2 & 80.3 &
9.1 & 31.0 & 53.9 & 69.1 &
7.6 & 24.1 & 40.1 & 49.2 \\

MonoDis (single)~\cite{simonelli2019disentangling} &
10.7 &
37.5 &
\underline{69.0} &
\textbf{85.7} &
- &
- &
- &
- &
- &
- &
- &
- 
\\

MonoDis (multi)~\cite{simonelli2020disentangling} & 
10.6 & 36.1 & 65.0 & 80.5 &
6.7 & 30.0 & 48.5 & 64.7 &
4.4 & 17.5 & 32.8 & 43.9 \\

FCOS3D~\cite{wang2021fcos3d}    &
\underline{15.3} & \underline{43.8} & 68.9 & 81.7 &
\underline{8.7} & \underline{30.3} & \underline{52.9} & \underline{67.1} & 
\underline{7.9} & \underline{25.0} & \underline{39.2} & \underline{47.1} \\

\midrule

DD3D &
\textbf{30.2} &
\textbf{59.7} &
\textbf{77.4} &
\underline{84.1} &
\textbf{18.7} &
\textbf{42.4} &
\textbf{61.9} &
\textbf{70.2} &
\textbf{15.7}  &
\textbf{32.6} &
\textbf{45.2} &
\textbf{50.0} 

\end{tabular}\\\vspace{0mm}
\caption{
\textbf{Detailed results on the nuScenes \textit{test} set.} We report AP metrics on \emph{Car}, \emph{Pedestrian}, and \emph{Bicycle} with varying thresholds on distance. The \underline{underline} denotes the second best published approach.
}
\label{table:nuscenes_test_detailed}
}
\end{table*}

%% file: sections/6_2_results_ablative.tex
\input{tables/depth_vs_det2d_v2}

Here we provide a detailed analysis of DD3D, focusing on the role of depth pre-training and comparison with our PL approach.  After pre-training the two models under various settings, we fine-tune them on KITTI-3D \emph{train} on the 3D detection task and report AP metrics on KITTI-3D \emph{val}. 

\subsubsection{Is depth-pretraining effective?}
\noindent{\textbf{Ablating large-scale depth pre-training.}} We first ablate the effect of dense depth pre-training on the DDAD15M dataset, with results reported in Table~\ref{table:kitti_3d_val_ablation}. When we omit the depth pre-training and directly fine-tune DD3D with the DLA-34 backbone on the detection task, we observe a $\mathbf{5.3\%}$ loss in \emph{Car Moderate  BEV AP}. In contrast, when we instead remove the initial COCO-pretraining (i.e. pre-training on DDAD15M \emph{from-scratch}), we observe a relatively small loss, i.e. $2.0\%$. For the larger backbone of V2-99, the effect of removing the depth pre-training is even more significant, i.e. $\mathbf{-10.7\%}$.



\noindent\textbf{Dense depth prediction as pre-training task.} To better quantify the effect of depth pre-training, we design a controlled experiment to further isolate its  effect (Table~\ref{table:pretraining_depth_vs_detection}).  Starting from a single set of initial parameters (COCO),  we consider two tasks for pre-training DD3D, 2D detection and dense depth prediction. The two pre-training phases use a \emph{common} set of images that are annotated with both 2D bounding box labels and sparse depth maps obtained by projecting Lidar pointcloud. To further ensure that the comparison is fair, we applied the same number of training steps and batch size ($15$K and $512$). The data used for this pre-training experiment is composed of $136571$ images from the nuScenes~\cite{caesar2020nuscenes} training set.


The experiment shows that, even with the smaller scale of pre-training data compared to DDAD15M ($137$K vs. $15$M), the dense depth pre-training yields a notable difference in the downstream 3D detection accuracy ($21.8\%$ vs. $20.5\%$ BEV AP on \emph{Car Moderate}).

\begin{figure}[h]
    \includegraphics[width=0.99\columnwidth]{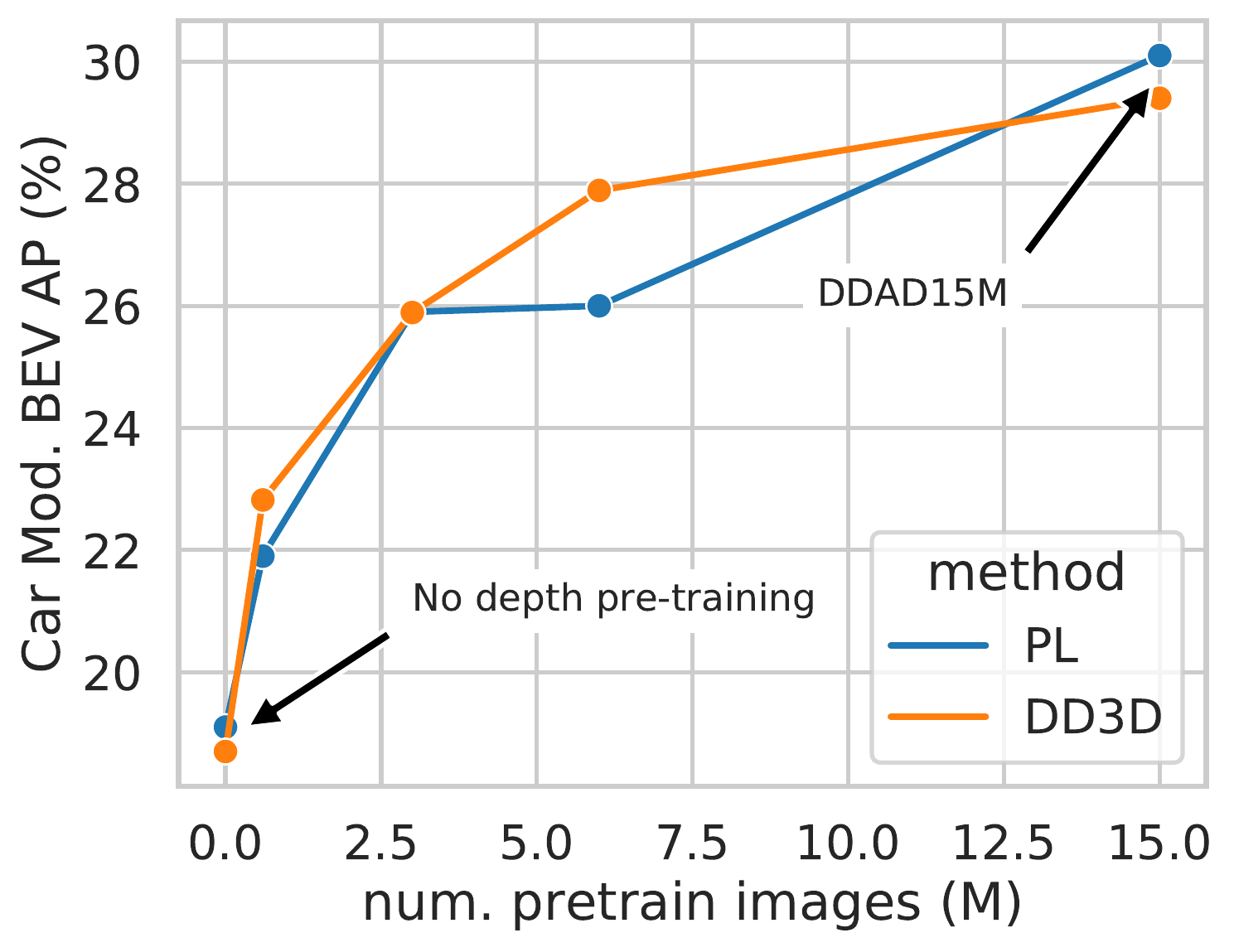}
    \caption{\textbf{DD3D and PL are pre-trained with varying sizes of depth data},  fine-tuned for 3D detection on the KITTI-3D \emph{train}, and evaluated on KITTI-3D \emph{val}. We pre-train the two methods on $4$ subsets of DDAD15M with $0.6$M, $3$M, $6$M, and $15$M images.}
    \label{fig:pretraining_vs_depth}
\end{figure}

\noindent\textbf{How does depth-pretraning scale?} We next investigate how the unsupervised depth pre-training scales with respect to the size of pre-training data (Figure~\ref{fig:pretraining_vs_depth}). For this experiment, we randomly subsample 1K, 5K, and 10K videos from DDAD15M to generate a total of 4 pre-training splits (including the complete dataset), that consist of 0.6M, 3M, 6M and 15M images respectively. Importantly, the downsampled splits contain fewer images as well as less diversity, as we downsample from the set of \emph{videos}. We pre-train both DD3D and the PackNet of PL on each  split, and subsequently train the detectors on KITTI-3D \emph{train}. We note that DD3D and PL perform similarly at each checkpoint, and continue to improve as more depth data is used in pre-training, at least up to 15M images. 


\subsubsection{The limitations of PL methods.}

\noindent\textbf{In-domain depth fine-tuning of PL.}
Recall that training our PL 3D detector entails fine-tuning of the depth network in the target domain (i.e. \emph{Eigen-clean}), after it is pre-trained on DDAD15M. We ablate the effect of the in-domain fine-tuning step in training the PL detector (Table~\ref{table:kitti_3d_val_ablation}). Note that in this experiment, the depth network (PackNet) is trained only on the pre-training domain (DDAD15M) and is directly applied on KITTI-3D without any adaptation. In this setting, we observe a significant loss in performance ($30.1\% \rightarrow 19.1\%$ BEV AP).  This indicates that in-domain fine-tuning of the depth network is crucial for PL-style 3D detectors. This poses a practical hurdle in using PL methods, since one has to curate a separate in-domain dataset specifically for fine-tuning the depth network. We argue that this is the main reason that PL methods are exclusively reported on KITTI-3D, which is accompanied by KITTI-Depth as a convenient source for in-domain fine-tuning. This is unnecessary for end-to-end detectors, as shown in Tables~\ref{table:kitti_3d_test} and~\ref{table:kitti_3d_val_ablation}.

\noindent\textbf{Limited generalizability of PL.} With the large-scale depth pre-training and in-domain fine-tuning, our PL detector offers excellent performance on KITTI-3D \emph{val} (Table~\ref{table:kitti_3d_val_ablation}). However, the gain from the depth pre-training is not transferred to the benchmark results (Table~\ref{table:kitti_3d_test}). While the loss in accuracy between KITTI-3D \emph{val} and the \emph{test} sets is consistent with other methods~\cite{simonelli2019disentangling, brazil2020kinematic, brazil2019m3d}, PL suffers particularly more ($30.1\% \rightarrow 18.6\%$ BEV AP), when compared to other methods including DD3D ($29.4\% \rightarrow 22.56\%$). This reveals a subtle issue in the generalization of PL that is not well understood yet. We argue that the in-domain fine-tuning overfits to some image statistics that cause the performance gap between KITTI-3D \emph{val} and KITTI-3D \emph{test}, particularly more than other methods.

%% file: tables/depth_vs_det2d_v2.tex
\begin{table}[h]
\centering
{
\footnotesize
\rowcolors{2}{lightgray}{white}
\begin{tabular}{l|ccc|ccc}
\toprule
& \multicolumn{3}{c}{BEV AP}  & \multicolumn{3}{c}{3D AP} \\ 
\multirow{-2}{*}{Pre-train task}& 
Car & 
Ped. &
Cyclist &
Car & Ped. & Cyclist \\
\midrule
COCO  & 
20.2 & 
7.8 &
\textbf{3.8} &
13.9 &
6.0 &
3.1
\\
\midrule
\quad + nusc-det & 
20.5 & 
7.9 &
\textbf{3.8} &
14.0 &
6.0 &
3.0 
\\
\quad + nusc-depth & 
\textbf{21.8} & 
\textbf{8.5} &
\textbf{3.8} &
\textbf{15.2} &
\textbf{6.6} &
\textbf{3.3}
\\
\bottomrule
\end{tabular}\\\vspace{0mm}
\caption{\textbf{Depth vs. 2D detection as pre-training task.} Starting from a common initial model (COCO), we pre-train DD3D using two different tasks using the same set of images. \emph{nusc-det} denotes 2D detection task, \emph{nusc-depth} denotes dense prediction task, both using nuScenes images. We report the accuracy on the KITTI-3D \emph{validation} set.
\label{table:pretraining_depth_vs_detection}
}
}
\end{table}

%% file: sections/7_conclusion.tex
\vspace{0.2cm}

We proposed \textit{DD3D}, an end-to-end single-stage 3D object detector that enjoys the benefit of Pseudo Lidar methods (i.e. scaling accuracy using large-scale depth data), but without its limitation (i.e. impractical training, issues in generalization). This is enabled by pre-training DD3D using a large-scale depth dataset, and fine-tuning on the target task end-to-end. DD3D achieves excellent accuracy on two challenging 3D detection benchmarks.  


%% file: sections/8_supplementary.tex
\section{Details of training DD3D and PL}
\label{sec:pretraining}
We provide the training details used for supervised monocular depth pre-training of both DD3D and PackNet.

\noindent\textbf{DD3D.} During pre-training, we use $512$ as a batch size, and train for $375$K steps until convergence. The learning rate starts at $0.02$, decayed by $0.1$ at the $305$K-th and $365$K-th steps. The size of the input images (and projected depth map) is 1600 $\times$ 900, and we resize them to 910 $\times$ 512. When resizing the depth maps, we preserve the sparse depth values by assigning all non-zero depth values to the nearest-neighbor pixel in the resized image space (note that this is different from naive nearest-neighbor interpolation, where the target depth value is assigned zero, if the nearest-neighbor pixel in the original image does not have depth value.) We observed that training converges after $30$ epochs. We use the Adam optimizer with $\beta = 0.99$.  For all supervised depth pre-training splits, we use an L1 loss between predicted depth and projective ground-truth depth.

 When training as 3D detectors, the learning rate starts at $0.002$, and is decayed by $0.1$, when the training reaches $85\%$ and $95\%$ of the entire duration. We use a batch size of $64$, and train for $25$K and $120$K steps for KITTI-3D and nuScenes, respectively. The $\mu_l$ and $\sigma_l$ are initialized as the mean and standard deviation of the depth of the 3D boxes that are associated with each FPN level, $\alpha_l$ as the stride size of the associated FPN level, and $c$ is fixed to $\frac{1}{500}$. The raw predictions are filtered by non-maxima suppression (NMS) using IoU criteria on 2D bounding boxes. For the nuScenes benchmark, to address duplicated detections in the overlapping frustums of adjacent cameras, an additional BEV-based NMS is applied across all $6$ synchronized images (i.e. a \emph{sample}) after converting the detected boxes to the global reference frame.

\noindent\textbf{PackNet.} When training PackNet~\cite{packnet}, the depth network of PL,  we use a batch size of 4 and a learning rate of $5\times10^{-5}$ with input resolution of  $640\times 480$. We use only front camera images of DDAD15M to pre-train PackNet, and train until convergence, and for 5 epochs over the KITTI \textit{Eigen-clean} split during fine-tuning. The PL detector is trained with a learning rate of $1\times10^{-4}$ for 100 epochs, decayed by $0.1$ after 40 and 80 epochs, respectively. For both networks we use the Adam~\cite{loshchilov2017decoupled} optimizer with $\beta_1=0.9$ and $\beta_2=0.999$. Both DD3D and PL are implemented using Pytorch~\cite{paszke2017automatic} and trained on 8 V100 GPUs.


\section{DD3D architecture details}
\label{sec:fcos_architecture}

\noindent\textbf{FPN}~\cite{lin2016fpn} is composed of a \emph{bottom-up} feed-forward CNN that computes feature maps with a subsampling factor of 2, and a \emph{top-down} network with lateral connections that recovers high-resolution features from low-resolution ones. The FPNs yield 5 levels of feature maps. DLA-34~\cite{yu2018deep} FPN yields three levels of feature maps (with strides of 8, 16, and 32). We add two lower resolution features (with strides of 64, 128) by applying two $3\times3$ 2D convs with stride of 2 (see Figure~\ref{fig:dd3d-arch}). V2-99~\cite{lee2019centermask} by default produces 4 levels of features (strides = 4, 8, 16, and 32), so only one additional conv is used to complete 5 levels feature maps. Note that the final resolution of FPN features derived from DLA-34 and V2-99 network are different, strides=8, 16, 32, 64, 128 for DLA-34, strides= 4, 8, 16, 32, 64 for V2-99. 

\noindent\textbf{2D detection head.} We closely follow the decoder architecture and loss formulation of~\cite{tian2019fcos}.
In addition, we adopt the positive instance sampling approach introduced in the updated arXiv version~\cite{wang2021fcos3d}.
Specifically, only the center-portion of the ground truth bounding box is used to assign positive samples in $\mathcal{L}_{reg}$ and $\mathcal{L}_\uppersub{3D}$ (Eq.~\ref{eq:loss_function}).

\section{Pseudo-Lidar 3D confidence head}
\label{sec:pl_detection_head}

Our PL 3D detector is based on~\cite{ma2020rethinking}, and outputs 3D bounding boxes with 3 heads, separated based on distance (i.e. near, medium and far). Following~\cite{simonelli2019disentangling,simonelli2020demystifying} we modify each head to output a 3D confidence, trained through the 3D bounding box loss. Specifically, each 3D box estimation head consists of 3 fully connected layers with dimensions $512 \longrightarrow 512 \longrightarrow 256 \longrightarrow  \left(\delta,\gamma\right)$, where $\delta$ denotes the bounding box parameters as described in~\cite{ma2020rethinking}, and $\gamma$ denotes the 3D bounding box confidence.

\section{The impact of data on Pseudo-Lidar depth and 3D detection accuracy}
\label{sec:supp_data_vs_3d_det}

\begin{figure}
    \includegraphics[width=0.99\columnwidth,trim={0mm 0mm 0mm 0mm},clip]{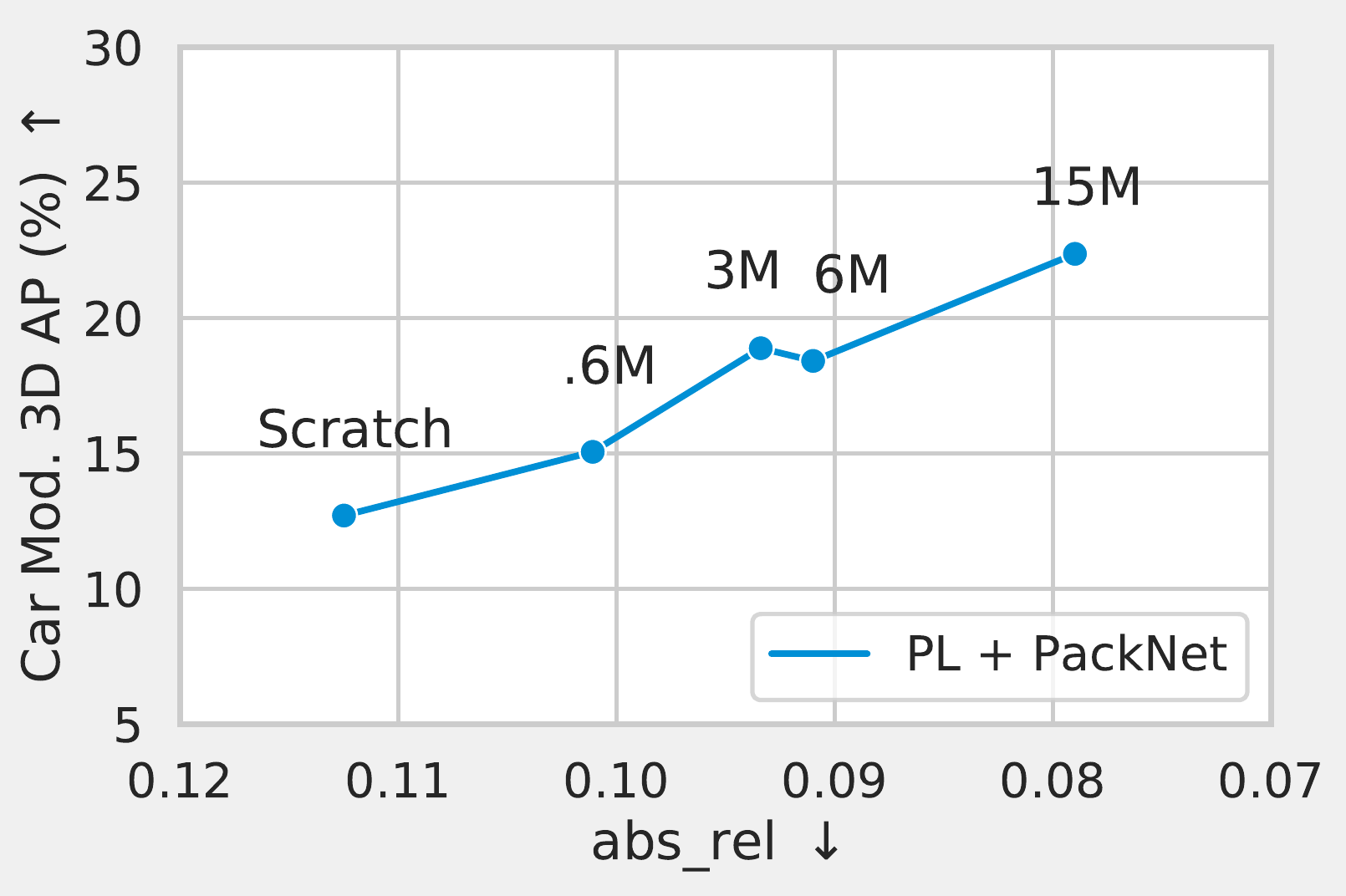}
    \caption{\textbf{We evaluate depth performance (abs\_rel) against PL 3D detection performance (Car Mod. 3D AP$|_{R_{40}}$) at each pre-training step.} All results are computed on the KITTI-3D \textit{validation} split.}
    \label{fig:3d_vs_depth}
\end{figure}

We evaluate depth quality against the 3D detection accuracy of the PL detector, with results shown in Figure~\ref{fig:3d_vs_depth}. Our results indicate an almost perfect linear relationship between depth quality as measured by the \textit{abs rel} metric and 3D detection accuracy for our PL-based detector.

\vspace{-5em}



